\pdfoutput=1

\documentclass[11pt,a4paper]{article}
\usepackage[hyperref]{emnlp2021}
\usepackage{times}
\usepackage{latexsym}
\usepackage{url}
\usepackage{color,soul}
\usepackage{amsmath,amssymb}
\usepackage{graphicx}
\usepackage{enumitem} 
\usepackage{tikz}
\usepackage{svg}
\usepackage{makecell}
\usepackage{multirow}
\usepackage{booktabs} 
\usepackage{subcaption}

\usepackage{microtype}

\title{What do writing features tell us about AI papers?}

\author{Zining Zhu$^{1,2}$, Bai Li$^{1,2}$, Yang Xu$^{1,2}$, Frank Rudzicz$^{1,2,3,4}$\\
$^1$Department of Computer Science, University of Toronto\\ $^2$Vector Institute for Artificial Intelligence \\
$^3$Li Ka Shing Knowledge Institute, Unity Health Toronto\\
$^4$Surgical Safety Technologies \\
  \texttt{\{zining, bai, yangxu, frank\}@cs.toronto.edu}}

\date{}

\begin{document}
\maketitle
\begin{abstract}

As the numbers of submissions to conferences grow quickly, the task of assessing the quality of academic papers automatically, convincingly, and with high accuracy attracts increasing attention. We argue that studying interpretable dimensions of these submissions could lead to scalable solutions. We extract a collection of writing features, and construct a suite of prediction tasks to assess the usefulness of these features in predicting citation counts and the publication of AI-related papers. Depending on the venues, the writing features can predict the conference vs. workshop appearance with F1 scores up to 60-90, sometimes even outperforming the content-based tf-idf features and RoBERTa. We show that the features describe writing style more than content. To further understand the results, we estimate the causal impact of the most indicative features. Our analysis on writing features provides a perspective to assessing and refining the writing of academic articles at scale.
\end{abstract}

\section{Introduction}
As technology continues to develop rapidly, conferences and journals have increased numbers of article submissions. There is increased criticism from multiple sides. On one side, authors criticize the randomness and subjectivity of peer reviews \citep{rogers-augenstein-2020-improve,Church2020}. On the other side, reviewers worry about the quality of papers, frequently raise concerns, and reject most submissions.

The tension reveals an underlying problem: it is hard to assess the quality of academic articles (1) automatically, (2) with high accuracy, and (3) in a way that is convincing to humans. An ideal evaluation system should meet all three criteria. Unfortunately, current systems can satisfy at most two of the three.
\begin{itemize}[nosep] 
    \item Human-based peer reviewing is the de-facto system. Thanks to the detailed reviewer comments, this system gives the most convincing results. While some components (e.g., paper - reviewer matching) can be automated, this system is largely run by humans.
    \item Automatic essay scoring (AES) systems are automatic, but they are developed and applied in the domain of student essay scoring, a completely different domain from scholarly articles. Transferring between domains limits the possible accuracy. 
    \item Deep neural network systems can automatically predict paper appearance with reasonable accuracy, but the granularities of their results are coarse, compared to feature-based AES systems.
\end{itemize}

Is that possible to build scalable systems that reach all three criteria simultaneously, to assist the human reviewers? In this paper, we argue that studying the interpretable properties of articles could lead to potential solutions.

Academic articles, despite some structural guidance or precedent, are fundamentally a form of discourse, where many features have been shown to have perlocutionary effects. For example, shorter sentences, and the use of certain rhetorical devices (e.g., amplification) can attract the readers' attention, and could boost the popularity of social media posts \citep{page2013stories}. These devices are especially popular among political discourse, and are also used widely \citep{janks1997critical,Browse2018,catalano2020critical}.

Looking from a discourse analysis perspective, a series of questions quickly emerge. Can the factors related to discourse, rather than the contents of an article, impact where the papers appear? If an article is written with shorter sentences, or written in more readable forms, will future researchers be more inclined to cite it? The query from these questions motivates us to study \textit{writing features}.

We compute a collection of writing features describing distinct discourse aspects of an article (e.g., the proportion of active voice in abstract), that are independent of content or topic (\S \ref{sec:features}). We construct a suite of prediction tasks using a dataset containing 945,674 published computer science papers (\S \ref{sec:data}). We find that the annual citation counts are hard to predict (\S \ref{subsec:predict_citations_regression}). However, the conference vs. workshop appearance of some top-tier venues can be predicted using writing features with F1 scores up to 60-90, sometimes even outperforming the content-based classifiers (\S \ref{subsec:predict_appearance}). With inter-venue and inter-category classification tasks, we illustrate that the writing features describe the styles instead of the contents (\S \ref{subsec:classify_between_venues}).
To further understand the results, we estimate the causal impacts of the most indicative features and study their empirical indications to paper publications (\S \ref{subsec:most_important_features}).

We release the writing features and the test suite\footnote{\url{https://github.com/SPOClab-ca/writing-features-AI-papers}}. Our work presents a new perspective towards building scalable and interpretable systems for assessing and refining academic articles.

\section{Related Work}
\paragraph{Reflections on Peer Review} Researchers have discussed how the peer review process could be improved \citep{Kelly2014,DeSilva2017}. Recently, the peer review process has been called into doubt \citep{Stelmakh2019Testing,Church2020}. \citet{Bharadhwaj2020} found that the presence on arXiv affects the acceptance decisions of ICLR papers, especially the borderline ones.
These reflections on peer review show the difficulty to assess the qualities of academic articles automatically, convincingly, and with high accuracy.

\paragraph{Automatic Essay Scoring}
A survey by \citet{ke2019AESsurvey} grouped features most used by state-of-the-art AES systems into ten categories: length, lexical, embedding, category-based, prompt-relevant, readability, syntactic, argumentation, semantics, and discourse. Each AES system used several of them. Some previous AES approaches also included deep learning \citep{wang-etal-2018-AES-RL,dong-zhang-2016-automatic} but, in this paper, we follow the feature engineering route. We expand this discussion in \S \ref{sec:features}.

\paragraph{Acceptance and citation link prediction}
Previous work automatically predicted the acceptance of academic papers using the gestalt \citep{Huang2018}, or the texts \citep{Yang2018AAPR,li-etal-2020-multi}. \citet{wang_paper_2021} predicted acceptance at the institutional level.
There are also methods to predict citation counts, using either engineered features \citep{yan_citation_2011} or BERT-based models \citet{van-dongen-etal-2020-schubert}. 
In this paper, we include the citation link prediction and the venue appearance (instead of acceptance, since most rejected papers are not publicly viewable) tasks in our evaluation suite. 

\paragraph{Causal analysis with text features} Causal reasoning on text features allows us to estimate the effects of the factors that are encoded in the texts \citep{kang-etal-2017-detecting,Wood-Doughty2018,egami2018make}, and to measure classifier performance attributable to textual attributes \citep{pryzant2018adsstyle} and lexicons \citep{pryzant2018deconfounded}. Recent works use causal reasoning to explain the model performance \citep{Feder2019}, and to reveal confoundings \citep{Vig2020causal,keith-etal-2020-text}. \citet{fytas2021makes} looks for interpretable factors contributing to paper acceptance. \citet{vincent-lamarre_textual_2021} studies the impact of some discourse features to AI paper acceptance. We use a simple model to estimate the causal effects of distinct writing features towards venue appearance (\S \ref{subsec:most_important_features}).

\section{Extracting Writing Features}
\label{sec:features}
We consider features depicting the quality of writing in an article, as in AES systems \citep{ke2019AESsurvey}, while avoiding content to the extent possible. Except for those defined on the whole article (e.g., title length), we compute each feature on the abstract and the bodytext respectively.

\paragraph{Citation counts features (1 label, 0 feature)}
\begin{itemize}[nosep]
    \item The annual inbound citations reflect the value of the articles, as perceived by other authors \citep{Hou2020}. Each article may receive a different number of inbound citations each year, but we count the \textit{normalized} number and refer to them as the annual citation counts henceforth.
    \item The total inbound citation counts. This is highly correlated to annual citation counts ($R=0.9472$, while the remaining features have at most $R=0.3$). The articles with high academic merit would be cited more annually and would have more accumulative citations, so we consider the total inbound citation count ``synonymous'' to the annual citations. Since we already have the annual counts, we exclude the total counts from the regression and classification experiments.
\end{itemize}

\paragraph{Article-metadata (3 features)}
\begin{itemize}[nosep]
    \item Title length, measured in the number of words. There are slight negative correlations between the title length and the total citations of articles \citep{letchford2015advantage}, but that can be explained by the scope of the content. The papers with shorter titles discuss more general topics, therefore may reach larger audiences.
    \item Number of authors.
    \item The outbound citations per word. This is the number of previous articles that the article that we analyze cite, normalized by the word counts. A more general article or a literature review tends to contain more outbound citations per word.
\end{itemize}

\paragraph{Article length (5 features)} These include the number of sections, the number of words and sentences in the abstract and body text. In subsequent analysis using 85 features, we include all 5. For those using 74 features (i.e., removing the effects of body text length), we include 2: the number of words and sentences in the abstract. Note that some features in other categories are also relevant to article length. For example, the number of grammatical errors in the body text, and the total number of outbound citations.

\paragraph{Sentence length (4 features)} The lengths of sentences are relevant to readability. Longer sentences are in general harder to read through, but strategic use of long and short sentences could contribute to the styles of writings \citep{strunk2007elements}. This category includes the mean and variance of lengths of each sentence. They are computed for the abstract and the bodytext respectively. Unless specified, each subsequent feature is likewise computed. 

\paragraph{Flesch Readability (4 features)} We use two scores, the Flesch readability ease (RE) \citep{flesch1948new} and the Flesch-Kincaid grade level (GL) \citep{kincaid1975derivation}, to describe the ease of reading.
The scores are computed as follows.
\begin{align*}
\text{RE} &= 206.835 - 1.015 \frac{W}{L} - 84.6 \frac{S}{W} \\
\text{GL} &= 0.39 \frac{W}{L} + 11.8 \frac{S}{W} - 15.59,
\end{align*}
where $S$, $W$, and $L$ are the syllable, word, and sentence counts, respectively. 

 Here, a higher syllable/word ratio $\frac{S}{W}$ correlates to the usage of more complex words, and a higher word/sentence ratio $\frac{W}{L}$ signals the use of relatively longer and more sophisticated sentences. 
Therefore, a higher readability ease (RE) value indicates the article being simpler, while a higher grade level (GL) indicates greater complexity. The RE and GL scores are used to refine the reading material for a potential audience, and assess the cognitive loads imposed by texts \citep{roberts2016readability,Kelly2017,wang2018csr,fakhoury2018effect}

\paragraph{Grammatical error count (2 features)} 
Grammatical errors have been used as a factor in AES systems to assess, e.g., language learners' writing abilities \citep{attali2006automated,shermis2003automated}.
We use a state-of-the-art grammar error correction model, open-sourced by Grammarly, GECToR \citep{omelianchuk-etal-2020-gector}. GECToR is a RoBERTa-based neural model\footnote{The authors also open-sourced other encoders like BERT and XLNet. RoBERTa was the default option.} which corrects the grammatical mistakes in a sequence tagging. 
We compute the number of ``grammar mistake'' tags recommended by GECToR in both the abstracts and the bodytexts, for each article. 

\paragraph{Lexical richness (10 features)}
Lexical richness has been used to analyze writing style \citep{smith2002stylistic}, vocabulary,  and writing skills \citep{LAUFER1995,gregori2015analysing}.
There are many methods (``indices'') to describe lexical richness \citep{Malvern2012}. In this paper, we use a method that is invariant to the article length: the moving-average type-token ratio (MATTR, \citet{covington2010cutting}). A higher MATTR value indicates a less repetitive usage of words.

To compute MATTR: First compute the type-token ratio (TTR): 
$$\text{TTR} = \frac{N. \text{types}}{N. \text{tokens}},$$
where the number of types is the number of distinct tokens. Then average over some fixed length windows (we use 5, 10, 20, 30, 40) to get the MATTR with 5 different window lengths. We use a Python library\footnote{\url{https://pypi.org/project/lexicalrichness}} to compute MATTR on the abstract and the bodytext of each article.

\paragraph{Part-of-speech constituency (28 features)} We compute the part-of-speech constituency. We use the 14 PoS tags (in English) given by SpaCy\footnote{\url{https://spacy.io}.}  The tags are: ADJ, ADV, ADP, AUX, CCONJ, DET, INTJ, NOUN, NUM, PART, PRON, PROPN, SPACE, VERB. The constituency of e.g., NOUN in abstract is computed by the percentange of occurrence of the NOUN tag in the abstract of an article. Various researchers consider part-of-speech to be an important signal related to the syntactical information encoded in the text \citep{Jurafsky2000SLP,UniversalDependencies2.5,Tenney2019EdgeProbing}. The choice of part-of-speech has been used as a marker for writing style \citep{campbell2003secret}, language fluency for foreign language learners \citep{alderson2005diagnosing}, and even cognitive capacity \citep{fraser15-JAD}.

\paragraph{Sentential surprisal (4 features)} This is computed by the average log perplexity of GPT-2 -- a pretrained uni-directional language model \citep{radford2019language} -- when reading through the first token of each sentence in an article. For an article consisting of $N$ sentences $\{s_1, ..., s_N\}$, the sentential surprisal is computed as:
\begin{align*}
    \text{SS} = \mathbb{E}_{i} \text{ log} P(s_{i+1}^{(0)}\,|\,s_{1..i}),
\end{align*}
where $s_{i}^{(j)}$ refers to the $j^{th}$ token in the $i^{th}$ sentence of the article.

We can interpret the sentential surprisal scores as follows. If there are large ``semantic gaps'' from sentence to sentence, then the overall difficulty for comprehension is increased, and a unidirectional language model will show higher perplexity. Table \ref{tab:surprisal_examples} presents two examples. Note that other factors can impact the perplexity values produced by the language models, including the word frequency and the choice of grammar. Note that both the word frequency \citep{shain2019large} and the grammatical forms \citep{gough1965grammatical} are relevant to the ease of understanding the text. 
Regardless, the ``semantic gaps'' are not necessarily causative to the sentential surprisal scores of the language models, but they are correlative \citep{goodkind-bicknell-2018-predictive}.

\begin{table}[t]
    \centering
    \resizebox{\linewidth}{!}{
    \begin{tabular}{l l}
    \toprule
        \textbf{Sentences} & \textbf{Surprisal} \\ \midrule 
        \makecell[l]{We propose a network based on BERT. \textul{We} describe \\the network as following.} & 90.194 \\ \midrule 
        \makecell[l]{We propose a network based on BERT. \textul{Recently}, \\deep neural networks are widely used.} & 92.861 \\ 
        \midrule 
        \bottomrule
    \end{tabular}}
    \caption{Two examples of sentences. The surprisal values are the log probability of the \textul{underlined} tokens, as computed by a pretrained GPT-2 \citep{radford2019language}. The former example follows a more natural flow of writing. In the second example, the background is written after the detailed activity on purpose, and it has a larger surprisal value.}
    \label{tab:surprisal_examples}
\end{table}

\paragraph{Rhetorical signal constituency (18 features)}
Academic papers, like other types of articles written with specific goals, contain many rhetorical devices. We use Rhetorical Structure Theory (RST) \citep{mann1987rhetorical} to quantitatively describe the rhetorical activities in the articles.

Recently, RST features have been used in developing some AES systems \citep{wang2019RST-ETS}. They are related to both the style and content of articles. For example, a theoretical article presenting an abstract idea may contain extensive \textit{elaboration} and \textit{explanation} signals, while a perspective article written to contribute to a debate likely contains more \textit{contrast} and \textit{comparison} signals. 

We parse the abstracts of the articles with a pretrained RST parser \citep{feng-hirst-2014-linear}, and count the proportion of each RST signal. 
Note that we only consider the abstracts due to time constraints. Building an RST parse tree for an abstract takes between 2 to 10 seconds. For a full article, this could take at least 5 minutes. Computing only the $\approx10^6$ articles in the Computer Science category of S2ORC \citep{lo-etal-2020-s2orc} would take at least 9 years of computation on available machines, which is unrealistic.

\paragraph{Active and passive voice proportions (6 features)} We count the proportions of active- and passive-voice sentences in both the abstract and the body text. To detect the voice of a sentence, we check the dependency tags of all of its tokens. If the sentence is in the active voice, its subject has a \textit{nsubj} tag. If it is in passive voice, its nominal subject has a \textit{nsubjpass} tag. If neither tag occurs in any token of the sentence, we label the sentence as in the ``Other'' voice. Table \ref{tab:voice_examples} shows some examples. 

\begin{table}[t]
    \centering
    \resizebox{\linewidth}{!}{
    \begin{tabular}{l l}
    \toprule
        \textbf{Sentence} & \textbf{Label} \\ \midrule 
        \makecell[l]{\textul{We} show that \textul{dropout} improves the performance of\\ neural networks on supervised learning tasks...} & Active  \\ \midrule 
        \makecell[l]{In the simplest case, each \textul{unit} is retained with\\a fixed probability p independent of...} & Passive \\ 
        \midrule 
        \makecell[l]{Applying dropout to a neural network amounts to \\sampling a ``thinned'' network from it.} & Other \\ \midrule 
        \bottomrule
    \end{tabular}}
    \caption{Several example sentences in different voices, taken from a highly cited paper \citep{JMLR:v15:srivastava14a}. The underlined words in the active and the passive voice sentence are tagged with the \texttt{nsubj} and \texttt{nsubjpass} dependency tags respectively.}
    \label{tab:voice_examples}
\end{table}

\begin{table*}[t]
    \centering
    \resizebox{\linewidth}{!}{
    \begin{tabular}{l |c| l l l l l | l l | l}
        \toprule 
        \multirow{2}{*}{\textbf{Venue Name}} & \multicolumn{6}{c|}{\textbf{Writing Features}} & \multicolumn{2}{c|}{\textbf{TF-IDF}} & \textbf{RoBERTa}\\
        & \hspace{1em}74 features\hspace{1em} & RST & Surprisal & Grammar & LexRich & Readability & Full text & Abstract & Abstract \\ \midrule 
        AAAI & $.755 (.028)$ & $+.001$ & $+.024$ & $+.010$ & $-.002$ & $+.009$ & $+.206^{**}$ & $+.203^{**}$ & $+.212^{**}$\\
        ACL & $.867 (.004)$ & $+.001 $ & $+.000$ & $+.001$ & $+.001$ & $+.001$ & $-.004$ &  $-.008^{*}$ & $-.015$ \\
        COLING & $.837 (.010)$ & $+.005$ & $+.005$ & $+.003$ & $+.003$ & $+.004$ & $+.049^{**}$ &  $+.051^{**}$ & $+.052^{**}$ \\
        CVPR & $.900 (.005)$ & $-.007$ & $-.006$ & $-.001$ & $-.006$ & $-.005$ & $-.052^{**}$ & $-.067^{**}$ & $-.070^{**}$ \\
        EMNLP & $.737 (.020)$ & $+.003$ & $+.014$ & $+.012$ & $+.022$ & $+.015$ & $+.159^{**}$ & $+.153^{**}$ & $+.102$ \\
        ICML & $.659 (.023)$ & $-.277^{**}$ & $-.042^{*}$ & $-.102$ & $-.262^{**}$ & $-.185$ & $+.333^{**}$ & $+.333^{**}$ & $+.300^{**}$ \\
        IJCAI & $.868 (.002)$ & $-.067^{**}$ & $-.066^{**}$ & $-.045^{**}$ & $+.075^{**}$ & $-.067^{**}$ & $-.029^{**}$ & $-.061^{**}$ & $-.076$ \\
        NAACL & $.757 (.019)$ & $+.016$ & $+.016$ & $+.011$ & $+.017^{*}$ & $+.016^{*}$ & $-.107^{**}$ & $-.128^{**}$ & $-.182$ \\
        NeurIPS & $.586 (.035)$ & $-.193^{**}$ & $-.039$ & $-.097$ & $-.212^{**}$ & $-.157$ & $+.031$ & $-.077^{*}$ & $-.110$\\ 
        \midrule \bottomrule
    \end{tabular}}
    \caption{F1 scores of the \texttt{C} vs. \texttt{W} classification results. The second column shows the mean and stdev of F1 scores using 74 writing features. The remaining columns show the values \textit{relative to} the second column. $*$ and $**$ indicate $p<.005$ and $p<.001$ respectively, both on 2-tailed $t$-test with $dof=10$, Bonferroni corrected. \\ 
    Interpretation: Usually, writing features do not classify as well as the content-based classifiers, but sometimes the difference is not significant (e.g., ACL). In CVPR and NAACL, writing features are even better.}
    \label{tab:top_or_not_venue_classify}
\end{table*}
\section{Data}
\label{sec:data}
We use the Semantic Scholar Open Research Corpus (S2ORC) \citep{lo-etal-2020-s2orc}. S2ORC is currently the largest publicly available collection of research articles, approximately 320 times larger than the ACL Anthology dataset \citep{Radev2009}.

There are 19 subjects in S2ORC. We use the Computer Science category, containing 4,305,658 articles. Among them, 994,434 articles contain full texts. We computed the features mentioned in section \ref{sec:features} and removed \texttt{nan} entries, resulting in 945,674 articles.

\paragraph{Citation profile}
Among these articles, 923,699 ($97.68\%$) have $\leq10$ annual incoming citations. 937,381 ($99.12\%$) have less than $\leq20$ annual incoming citations. For simplicity, we use the term ``citation'' to refer to \textit{incoming} citations (i.e., how many times an article is cited), and do not abbreviate the outbound citation counts (i.e., how many times it cites others) in the rest of this article.
On average, each article is cited $1.59$ (std=$13.5$) times per year. As shown in Figure \ref{fig:citation_profile} (in Appendix), the number of articles decreases exponentially as the annual citation counts increase.

\paragraph{Article categories} The S2ORC dataset provides noisy labeling of the venue and journal occurrences for each article. For simplicity, we refer to ``venue or journal'' as ``venue''. We filter some venue names in various categories with keyword-based regex matching. In NLP, ML, AI, and CV, there are $6,681$ venues/journals containing about 465k CompSci articles. Table \ref{tab:articles_by_category} (in Appendix) shows the details.

\paragraph{Venue labels} Additionally, we mark each venue of our selected categories with a binary \texttt{C} or \texttt{W} label. In general, a \texttt{C} label stands for top-tier conferences, where a \texttt{W} label stands for workshops and general arXiv papers. That said, workshops with high impacts (as shown in the Google Scholar Metrics and guide2research), including SemEval, *SEM, and Rep4NLP, are labeled as \texttt{C} as well.
Table \ref{tab:top_or_not_label_examples} (in Appendix) contains some examples of labeling. Table \ref{tab:venue_numbers} shows the numbers of articles in each venue. Specifically, we consider 9 top-tier venues:
\begin{itemize}[nosep]
    \item NLP: ACL, NAACL, EMNLP, COLING
    \item Machine learning: NeurIPS, ICML
    \item Artificial intelligence: AAAI, IJCAI
    \item Computer vision: CVPR
\end{itemize}
These 9 venues\footnote{Additionally, we label a robotics venue: ICRA. The regex matched imbalance numbers of \texttt{C} vs. \texttt{W} papers here, so we exclude ICRA in \ref{subsec:predict_appearance}. We still include robotics in \ref{subsec:classify_between_venues}.} (including their workshops) contain 12,747 articles. Among the top-tier papers, the mean annual citation count is $7.28$ (std=$67.8$). Their citation profile also obeys a exponential distribution (Figure \ref{fig:citation_profile} in Appendix).

\begin{figure*}[t]
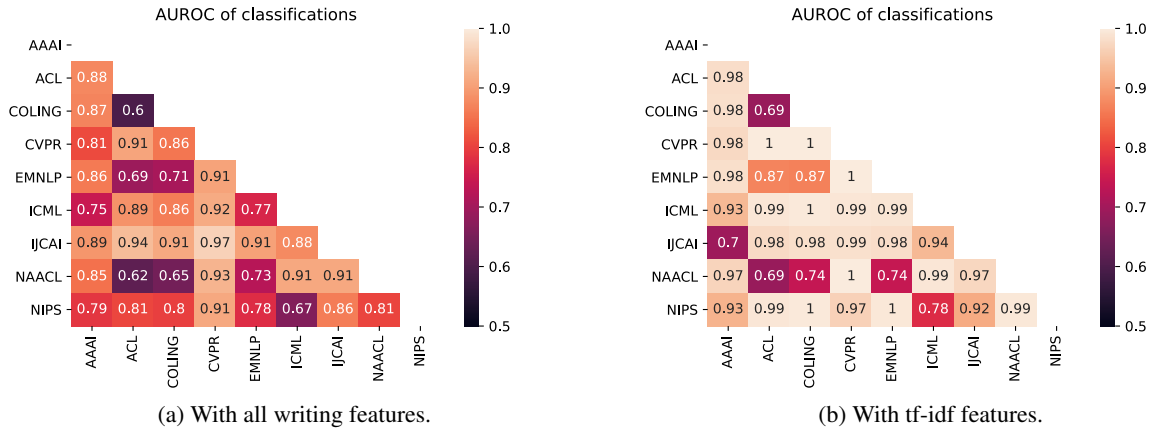

\begin{subfigure}[b]{.49\linewidth}
    \includesvg[width=\linewidth]{fig/venue_auroc_classifications.svg}
    \caption{With all writing features.}
\end{subfigure}
\hspace{1em}
\begin{subfigure}[b]{.49\linewidth}
    \includesvg[width=\linewidth]{fig/venue_auroc_classifications_tfidf.svg}
    \caption{With tf-idf features.}
\end{subfigure}
    \caption{The AUROC of inter-venue classifications. The venues in the same categories (e.g., COLING and ACL) are harder to tell apart than other venues, using either the writing features or tf-idf features.}
    \label{fig:heatmap_venue_classification}
\end{figure*}
\begin{figure*}
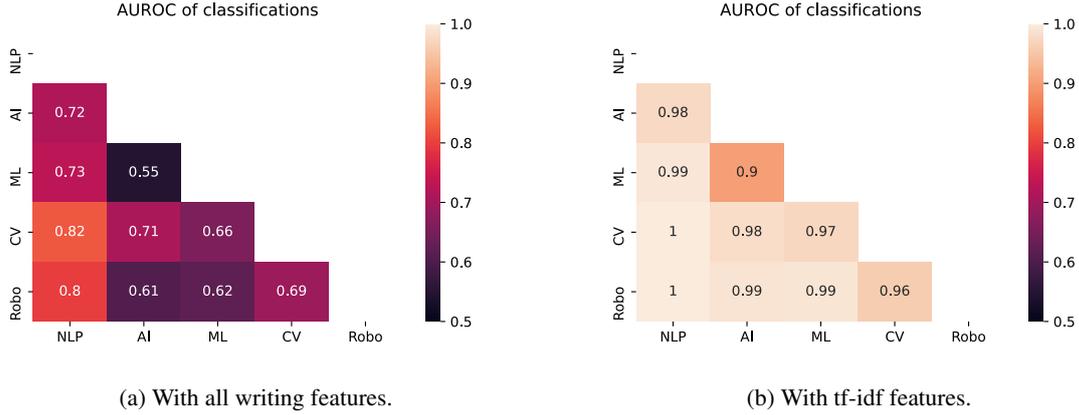

    \centering
    \begin{subfigure}[b]{.49\linewidth}
    \includesvg[width=\linewidth]{fig/category_auroc_classifications.svg}
    \caption{With all writing features.}
    \end{subfigure}
    \begin{subfigure}[b]{.49\linewidth}
    \includesvg[width=\linewidth]{fig/category_auroc_classifications_tfidf.svg}
    \caption{With tf-idf features.}
    \end{subfigure}
    \caption{The AUROC of inter-category classifications. The content-based tf-idf features can better predict the categories of the articles than the writing features ($p<.01$ for all categories, 2-tailed $t$ test with $dof=8$).}
    \label{fig:heatmap_category_classification}
\end{figure*}

\section{Experiments}

\subsection{Citation counts are hard to predict}
\label{subsec:predict_citations_regression}
The annual citation counts is perhaps the most objective quantity related to the impact of articles. It turns out that the current writing features, no matter when using together or in most combinations, cannot predict annual citation counts better than a trivial baseline (the mean annual citation counts of training data). 
The annual citation counts seem more relevant to the contents, instead of the writing styles. The content-based features (i.e., tf-idf) produce MSEs with smaller variances, but they do not significantly outperform the baselines either. We include the detail in Appendix \ref{subsec:appendix-regression-detail}.

\subsection{Writing features can predict conference vs. workshop appearance}
\label{subsec:predict_appearance}
The conference (\texttt{C}) versus workshop (\texttt{W}) discrepancy is an interesting prediction task because many workshops affiliate with conferences. This way, \texttt{C} and \texttt{W} papers are usually about closely relevant topics but written in different styles (due to e.g., page limit requirements).

We let classifier models predict the \texttt{C} vs \texttt{W} appearance on the venues using the writing features, tf-idf features, and RoBERTa \citep{Liu2019roberta}. The experiment detail is in Appendix \ref{subsec:appendix-classification-detail}.

Table \ref{tab:top_or_not_venue_classify} shows our results. In general, writing features do not classify the article appearance as good as the content-based tf-idf features (e.g., in AAAI, COLING, EMNLP), but sometimes the difference is not significant (e.g., ACL, NeurIPS). Specifically, in CVPR and NAACL, using 74 writing features\footnote{We exclude 11 features here: 3 describing article lengths (since we don't want the classifiers to rely on ``shortcuts'' including the different page limits of long/short papers, and conferences/workshops), and 8 about MATTR (we keep only window size 10, and discard those window sizes 5, 20, 30, 40, for abstract and bodytext, since MATTR of different window sizes are highly correlative to each other, and we want to avoid multicollinearity). We find that dropping these features almost never makes a statistically significant difference. The ablation study detail is listed in Table \ref{tab:top_or_not_features_ablation} (in Appendix).} significantly outperforms the tf-idf features. These illustrate the usefulness of the writing features.

Should we use all features or a partial set? Since there is no clear evidence suggesting otherwise, we proceed with the collection of 74 features in subsequent analysis to get broad coverage. As we will elaborate in \S \ref{subsec:most_important_features}, the writing features are mutually dependent, so classifying with a partial set can still get comparable results.

\subsection{Writing features describe style more than content}
\label{subsec:classify_between_venues}
The aforementioned \texttt{C} vs. \texttt{W} classification is an intra-venue prediction task. Within each venue, one can argue that the articles follow certain styles. To examine if the writing features describe these styles, we run pairwise classifications between (1) each pair of conferences and (2) each pair of categories, using writing features and tf-idf features.

The first observation is, venues in the same category are harder to tell apart. As shown in Figure \ref{fig:heatmap_venue_classification}, the AUROC scores between e.g., ACL and NAACL are lower than that of e.g., ACL and CVPR. However, even the lowest classification performance\footnote{ACL vs. COLING, at 0.60 AUROC with writing features and 0.69 AUROC with tf-idf features.} is much higher than what we would expect from random guessing (0.50 AUROC). This shows that the styles of ACL COLING papers, while similar, are still slightly different. A potential reason is that ACL and COLING have slightly different tracks, and authors have slightly different preferences when writing ACL vs. COLING papers.

Second, in almost all pairwise classification tests, tf-idf features outperform the writing features\footnote{$p<.01$ except NAACL vs EMNLP ($p=.11$), on 2-tailed $t$ test with $dof=8$.}. This may result from the vocabulary difference across venues -- we don't usually see words like ``discourse'' in computer vision papers.

Third, while it is relatively easy to distinguish papers between representative venues, it is harder to tell papers apart across categories using writing features, as illustrated by smaller AUROC scores in Figure \ref{fig:heatmap_category_classification}(a) than Figure \ref{fig:heatmap_venue_classification}(a).
However, this does not apply to the content-based classifications -- the use of words contains sufficient information to distinguish between the categories, as is captured by tf-idf features. 
We consider this a result of the \textit{stylistic diversity}: The articles can be written in diverse styles regardless of the category. Both an NLP and an AI paper can be written in short sentences and readable forms, but their contents are different.

\begin{table*}
    \centering
    \resizebox{.9\linewidth}{!}{
    \begin{tabular}{c l l l l}
    \toprule 
        \textbf{Venue} & \textbf{Features} & \textbf{Spearman R} & \textbf{ATE} & \textbf{Interpretation} \\ \midrule 
        \multirow{5}{*}{ACL} & \texttt{flesch\_kincaid\_grade\_level\_bodytext} & $-0.05$ & $+0.05$ & Ambiguous\\
        & \texttt{grammar\_errors\_abstract} & $-0.09^{**}$ & $-0.01$ & \texttt{W} papers are larger\\
        & \texttt{surprisal\_abstract\_std} & $-0.01$ & $+0.00$ & Ambiguous \\
        & \texttt{title\_word\_length} & $-0.09^{**}$ & $-0.01$ & \texttt{W} papers are larger\\
        & \texttt{voice\_bodytext\_active} & $+0.09^{**}$ & $+0.15$ & \texttt{C} papers are larger\\ \midrule
        \multirow{5}{*}{EMNLP} & \texttt{outbound\_citations\_per\_word} & $-0.17^{**}$ & $+67.6$ & Ambiguous \\
        & \texttt{n\_author} & $-0.17^{**}$ & $-0.05$ & \texttt{W} papers are larger \\
        & \texttt{grammar\_errors\_abstract} & $-0.18^{**}$ & $+0.01$ & \texttt{W} papers are larger \\
        & \texttt{n\_outbound\_citations} & $-0.09$ & $+0.09$ & Ambiguous \\
        & \texttt{abstract\_word\_counts} & $-0.16^{**}$ & $+0.00$ & \texttt{W} papers are larger \\ \midrule \bottomrule
    \end{tabular}}
    \caption{The most important 5 writing features for classifying \texttt{C} vs. \texttt{W} appearance, their Spearman R, and their estimated average treatment effects (ATE), taking ACL and EMNLP as examples. $*$ and $**$ indicate $p<.005$ and $p<.001$ respectively (Bonferroni corrected). For Spearman R, the $p$ value indicates the likelihood the feature and target come from identical distributions. For ATE, the $p$ value is computed by \textit{doWhy}'s default bootstrapping test.}
    \label{tab:important_features}
\end{table*}

\subsection{Writing features describe interpretable characteristics of venue appearance}
\label{subsec:most_important_features}

Here we attempt to further understand the classification results by studying the top 5 features identified by the best classification models in \S \ref{subsec:predict_appearance}. We compute the Spearman correlation (to the appearance) and the Average Treatment Effect (ATE) for each feature.

\begin{figure}[t]
    \centering
    \includegraphics[width=.5\linewidth]{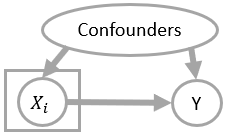}
    \caption{A simple causal model for estimating the causal effects of the writing features ($\{X_j\}$) towards the paper appearance ($Y=1$ for conference, and $Y=0$ otherwise).}
    \label{fig:causal_model}
\end{figure}

The ATE of a feature $X_j$ to the target $Y$ is the expected partial derivative $\frac{\partial \mathbb{E}Y}{\partial X_j}$. 
We estimate the ATE values with \textit{doWhy} \citep{dowhy}'s ``backdoor.linear\_regression'' algorithm, using a simple causal model. As shown in Figure \ref{fig:causal_model}, we assume each of the 74 writing features is causally related to the target while being independent of each other. 

Spearman R and ATE provide two aspects of understanding the features independently. If both scores have the same signs, we consider this feature to be \textit{indicative} of the target. Table \ref{tab:important_features} shows the indicative features of ACL and EMNLP as examples. The full table (Table \ref{tab:important_features_full}) is included in Appendix.

\texttt{grammar\_errors\_abstract} has both negative SpearmanR and ATE values, indicating that \texttt{C} papers are more likely to have fewer grammar errors than \texttt{W} ones. In addition, grammar error counts are identified as a ``top-5'' feature in AAAI (abstract, 3rd), CVPR (bodytext, 1st), EMNLP (abstract, 3rd), ICML (bodytext, 4th), IJCAI (bodytext, 4th), and NAACL (bodytext, 2nd). It has negative Spearman R values except for CVPR \footnote{$R=.03$ but $p=.14$. Also: $ATE=0$}. Reducing the grammar error counts may be beneficial for having a paper appear in top-tier conferences.

However, not all features have polarities that are as obvious. The \texttt{flesch\_kincaid\_ grade\_level} of bodytext is also identified as a top-5 feature frequently, but its polarity is more ambiguous. Its SpearmanR is negative in AAAI ($-0.08^{*}$) and ACL ($-0.05^{*}$), but positive in NeurIPS ($+0.05$), but in all three scenarios, its ATE estimates take the exact opposite signs. The ambiguous numbers prevent conclusive results about their polarities. In other words, the readability (Flesch-Kincaid grade level) of bodytext does not itself predict conference or venue publications.

Another example is \texttt{title\_word\_length}. The \texttt{C} papers are correlated to shorter title word lengths (4th in ACL, 1st in COLING), echoing the findings of \citet{letchford2015advantage}. In contrast to their explanation that papers with shorter titles are more readable, our study supports an alternative explanation: that workshop papers are more specialized, leading to their longer titles than the more general conference papers.

Only $22\%$ (10 out of the 45) ``top-5 features'' have significant estimated causal effects ($p<0.05$ for estimated ATE). This indicates that it is hard to single out a writing feature, manipulate its value (e.g., by modifying the writing style), and facilitate paper publication at a conference. We think the reason is that the writing features are \textit{dependent} -- any stylistic change will affect multiple writing features simultaneously. If an article is written in a more readable style, its sentences are likely shorter, its lexical richness may be smaller, and its part-of-speech constituency may change correspondingly. The inter-dependency of features also explains why using partial collections of features can usually lead to comparable performance as the full set of features (e.g., as shown in Table \ref{tab:top_or_not_venue_classify}). In the future, our analysis can be extended by e.g., grouping the features into mutually independent feature sets.

\section{Discussion}
\paragraph{Writing features are more than just writing.} We try to ensure these features reflect the writing, and that they do not explicitly correlate to content -- but they might do by hidden confounds. For example, if an article compared its proposed model to many other works in its experiments, its normalized outbound citation count could be large. However, a highly impactful paper, especially one opening up a novel direction, may not need to cite many other papers (e.g., backpropagation \citep{rumelhart1985learning} cites 11 other papers and is cited over 27,000 times). In short, there are various reasons to expect correlation or a lack thereof, the resolution of which we leave to future work.

\paragraph{Good papers are more than well-``written''.} Many highly impactful articles have appeared in workshops before. On one hand, studying the interpretable features inspires us to refine the writing e.g., by reducing the grammatical errors and paying attention to the readability. On the other, we should never ignore the intrinsic, academic quality of articles and their inspirations to future researchers.

\paragraph{Discourse features are text markers.} We can use the writing features in different scenarios, including the debates in online forums, and the interactions during the author response periods (e.g., similar to \citet{gao-etal-2019-rebuttal}). Further, discourse features can be text markers (resembling bio-markers), with which we can quantify and even factor out the undesired impacts to the readers. We can also use interpretable text features to diagnose model predictions, identifying some potential ``right for the wrong reason'' phenomena \citep{ThomasMcCoy2019}.

\section{Conclusion}
In this paper, we study the academic articles through a collection of \textit{writing features} that describe the interpretable dimensions of their styles without explicitly describing their contents. We compile a suite of prediction tasks to validate the effectiveness of these features. These writing features can predict the conference versus workshop appearance of some top-tier venues, sometimes outperforming the content-based tf-idf features and even RoBERTa. Examining the causal impacts of the indicative features leads to practical discussions about paper quality. Our analysis show a perspective towards automatically assessing and refining the writing of academic articles.

\bibliography{main}
\bibliographystyle{acl_natbib}

\newpage 

\appendix

\section{Appendices}

\subsection{Exploratory tables}

\begin{table}[h]
    \centering
    \begin{tabular}{l r r r}
        \toprule 
        \multirow{2}{*}{\textbf{Category}} & \multirow{2}{*}{\textbf{N. articles}} & \multicolumn{2}{c}{\textbf{N. articles by label}} \\ \cline{3-4}
        & & \texttt{C} & \texttt{W} \\ \midrule 
        AI & 23,642 & 3,493 & 20,149 \\
        CV & 29,881 & 20,029 & 9,852 \\
        ML & 12,628 & 6,196 & 6,432 \\
        NLP & 23,827 & 14,164 & 9,663 \\
        Robotics & 15,634 & 3,311 & 12,323 \\
        Speech & 8,123 & 6,576 & 1,547 \\
        Others & 831,939 & -- & -- \\ \midrule \bottomrule 
    \end{tabular}
    \caption{Article counts of AI-related papers by category. Note that some venues occur in multiple categories.}
    \label{tab:articles_by_category}
\end{table}

\begin{table}[h]
    \centering
    \resizebox{\linewidth}{!}{
    \begin{tabular}{l c r}
        \toprule 
        \textbf{Venue} & \textbf{Label} & \textbf{N. articles} \\ \midrule
        ACL & \texttt{C} & 2,338 \\ \midrule 
        \makecell[l]{Annual Meeting Of The Association \\ For Computational Linguistics} & \texttt{C} & 1,434 \\ \midrule 
        \makecell[l]{2019 IEEE/CVF Conference on \\Computer Vision and \\Pattern Recognition (CVPR)} & \texttt{C} & 1,274 \\ \midrule 
        Rep4NLP@ACL & \texttt{C} & 40 \\ \midrule 
        SemEval@NAACL-HLT & \texttt{C} & 698 \\ 
        \midrule 
        BioNLP@ACL & \texttt{W} & 73 \\ \midrule 
        WASSA@EMNLP & \texttt{W} & 84 \\ \midrule 
        AAAI Spring Symposia & \texttt{W} & 199 \\ \midrule 
        arXiv: Learning & \texttt{W} & 298 \\ \midrule 
        arXiv: Machine Learning & \texttt{W} & 258 \\ \midrule 
        \bottomrule
    \end{tabular}}
    \caption{Examples of venue names and their labels.}
    \label{tab:top_or_not_label_examples}
\end{table}

\begin{table}[h]
    \centering
    \resizebox{\linewidth}{!}{
    \begin{tabular}{l| r r r}
        \toprule 
        \multirow{2}{*}{\textbf{Venue Name}} & \multirow{2}{*}{\textbf{N. articles}} & \multicolumn{2}{c}{\textbf{N. articles by label}} \\ \cline{3-4}
         & & \hspace{3em}\texttt{C} & \texttt{W} \\ \midrule 
        AAAI & 624 & 395 & 229 \\
        ACL & 2,836 & 2,175 & 661 \\
        COLING & 1,860 & 1,353 & 507 \\
        CVPR  & 3,495 & 2,824 & 671 \\
        EMNLP & 714 & 437 & 277 \\
        ICML & 930 & 396 & 534 \\
        ICRA & 703 & 662 & 41 \\
        IJCAI  & 632 & 423 & 209 \\
        NAACL  & 2,142 & 1,354 & 788 \\
        NeurIPS  & 930 & 396 & 534 \\ \midrule \bottomrule 
    \end{tabular}}
    \caption{Number of \texttt{C} and \texttt{W} articles of each venue. The arXiv papers of the corresponding sections are included as \texttt{W} papers. For example, \texttt{cs.Learning} and \texttt{cs.ML} are included in the \texttt{W} portions of ICML and NeurIPS.}
    \label{tab:venue_numbers}
\end{table}

\begin{table*}[t]
    \centering
    \resizebox{\linewidth}{!}{
    \begin{tabular}{l |r r r r r r | l | r r}
        \toprule 
        \multirow{2}{*}{\textbf{Venue Name}} & \multicolumn{6}{c|}{\textbf{Writing Features}} & \multirow{2}{*}{\hspace{2em}\textbf{Baseline}} & \multicolumn{2}{c}{\textbf{TF-IDF}}\\
        & 74 features & RST & Surprisal & Grammar & LexRich & Readability &  & Full text & Abstract \\ \midrule 
        AAAI & $+8.61$ & $+0.09$ & $+0.69$ & $+0.27$ & $+1.16$ & $+0.61$ & \hspace{1.5em}$20.77 (27)$ & $0.07 (.02)$ & $0.07 (.01)$ \\
        ACL & $+6.27$ & $+0.48$ & $+0.11$ & $+0.10$ & $+0.34$ & $+2.89$ & \hspace{1em}$389.76 (636)$ & $0.15 (.01)$ & $0.17 (.01)$\\
        COLING & $+248.87$ & $+3.51$ & $+0.05$ & $+0.32$ & $+0.62$ & $+368.18$ & \hspace{1em}$437.76 (1006)$ & $0.15(.02)$ & $0.16(.01)$\\
        CVPR & $-6.11$ & $+239.82$ & $+9.62$ & $+6.90$ & $+22.24$ & $+12.35$ & $15273.45 (24710)$ & $0.17(.01)$ & $0.19(.01)$ \\
        EMNLP & $+55211.48$ & $+8.12$ & $+4.66$ & $+42.06$ & $+12.11$ & $+458.65$ & \hspace{0.5em}$1194.59 (2788)$ & $0.15(.02)$ & $0.17(.03)$ \\
        ICML & $+45.13$ & $+6.93$ & $+37.59$ & $+9.77$ & $+988.27$ & $+86.97$ & \hspace{0.5em}$1279.15 (1200)$ & $0.02(.01)$ & $0.02(.02)$\\
        IJCAI & $+8.84$ & $+1.37$ & $+0.81$ & $+1.20$ & $+3.77$ & $+1.97$ & \hspace{1.5em}$23.77 (25)$ & $0.16(.01)$ & $0.22 (.04)$\\
        NAACL & $+18.04$ & $+2.22$ & $+0.83$ & $+0.99$ & $+0.08$ & $+195.92$ & \hspace{1em}$420.34 (855)$ & $0.22(.01)$ & $0.22 (.01)$\\
        NeurIPS & $+78.25$ & $+4.64$ & $+6.37$ & $+39.81$ & $-2.87$ & $+76.99$ & \hspace{0.5em}$3305.99 (5216)$ & $0.20(.01)$ & $0.23 (.01)$\\ 
        \midrule \bottomrule
    \end{tabular}}
    \caption{Mean Squared Errors (MSEs) of the regression results predicting annual citation counts. The ``Baseline'' and ``TF-IDF'' columns show the mean and std of MSE. Other columns show MSEs \textit{relative to} the ``Baseline'' column. No values are significantly different from the ``Baseline'' column (partly due to the large variation of the writing features and the baselines), on 2-tailed $t$-test with $dof=10$, Bonferroni corrected. However, the MSEs of tf-idf features have much smaller variances. \\ Interpretation: The annual citation counts cannot be easily predicted by the writing features. On the other hand, the content-based features can predict annual citation counts with small mean squared errors.
    \label{tab:regression_results_by_venue}}
\end{table*}

\begin{figure*}
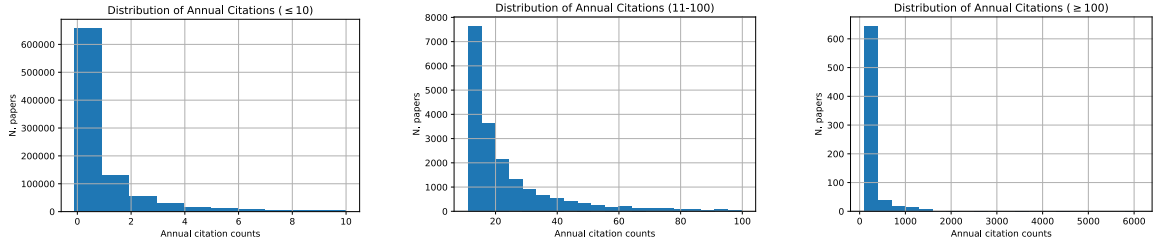
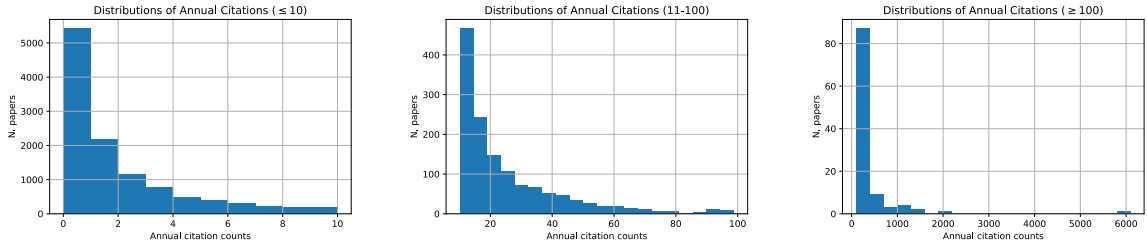

    \centering
    \begin{subfigure}[h]{\linewidth}
    \includesvg[width=.32\linewidth]{fig/citation_histogram_leq_10.svg}
    \includesvg[width=.32\linewidth]{fig/citation_histogram_11_100.svg}
    \includesvg[width=.32\linewidth]{fig/citation_histogram_geq_100.svg}
    \caption{The histogram plots of annual citations profiles of all CompSci papers.}
    \end{subfigure}
    \begin{subfigure}[h]{\linewidth}
    \includesvg[width=.32\linewidth]{fig/ai_citation_profile_0_10.svg}
    \includesvg[width=.32\linewidth]{fig/ai_citation_profile_11_100.svg}
    \includesvg[width=.32\linewidth]{fig/ai_citation_profile_100plus.svg}
    \caption{The histogram plots of annual citations profiles of the AI-related papers.}
    \end{subfigure}
    \caption{The histograms of annual citation profiles. The number of articles decreases exponentially as the annual citation counts increase. \label{fig:citation_profile}}
\end{figure*}

\subsection{Regression experiments details}
\label{subsec:appendix-regression-detail}
Following are the models used for predicting the annual citation counts in Table \ref{tab:regression_results_by_venue}:
\begin{itemize}[nosep]
    \item SVM (\texttt{LinearSVR}) with \{L1, L2\} loss, and $C=\{0.5, 1.0, 2.0\}$ regularization.
    \item LinearRegression, with and without fitting the intercept. When fitting the intercept, with and without normalization.
    \item ExtraTreesRegressor, with random state 0 and \{16, 32, 64, 128\} estimators.
    \item RandomForestRegressor, with random state 0 and \{50, 100, 200\} estimators.
    \item Gradient boosting, with random state 0 and maximum depths \{2,3,4,5\}
    \item Multiple Layers Perceptrons of various hidden sizes: [10], [20], [40], [80], [10,10], [20,20], and [40,40]
\end{itemize}
On all 9 venues, running 6 folds of experiments, including sweeping through all above models, takes 2-3 minutes on a desktop machine. The regression on all Computer Science articles (Table \ref{tab:regression_results_all_articles}) uses default MLPRegressor, and takes much longer -- around one hour for all folds.

\begin{table}[t]
    \centering
    \resizebox{.8\linewidth}{!}{
    \begin{tabular}{l c}
    \toprule
    \textbf{Configuration} & \textbf{MAE (stdev)} \\ \midrule
        \textit{All features} & 2.25 (0.05) \\ \midrule 
        \multicolumn{2}{l}{\textit{Rank by $|$Spearman R$|$}} \\
        Top 10 features & 1.98 (0.04) \\
        Top 20 features & 2.00 (0.03) \\
        Top 40 features & 2.16 (0.07) \\ \midrule 
        \multicolumn{2}{l}{\textit{Article features}} \\
        Part-of-speech & 2.03 (0.04) \\
        Rhetorical features & 2.02 (0.06) \\
        Sentential surprisal & 2.00 (0.05) \\
        Lexical richness & 1.98 (0.03) \\
        Grammar features & 1.97 (0.03) \\
        Sentence lengths & 1.97 (0.04) \\
        Readability & 1.96 (0.03) \\
        Voice ratio features & 1.96 (0.04) \\
        Article metadata & 1.93 (0.02)* \\
        Article lengths & 1.92 (0.08) \\ \midrule 
        \textit{Baseline}: Mean of train data & 1.99 (0.02) \\
    \midrule \bottomrule 
    \end{tabular}}
    \caption{Annual citation prediction results on all CompSci papers, using different features. The article-metadata features (title length, num. authors, and num. outbound citations) can predict significantly better than the trivial baseline ($p<.005$ for 2-tailed $t$-test, dof=8, Bonferroni corrected). Other features, while showing some signs of usefulness, are not as predictive.\\ Interpretation: The annual citation count prediction task requires more information than what is described by writing features.}
    \label{tab:regression_results_all_articles}
\end{table}

\begin{table*}
    \centering
    \resizebox{.7\linewidth}{!}{
    \begin{tabular}{l|l l l l l}
        \toprule
        \multirow{2}{*}{\textbf{Venue Name}} & \multicolumn{5}{c}{\textbf{Writing Features}} \\
        & All 85 features & All$\backslash$length & 74 features\hspace{1em} & Part-of-speech & Voice \\ \midrule
        AAAI & \hspace{1em}$+.010$ & $+.013$ & $.755 (.028)$ & $+.014$ & $+.008$ \\
        ACL & \hspace{1em}$+.002$ & $-.004$ & $.867 (.004)$ & $+.000$ & $+.001$ \\
        COLING & \hspace{1em}$-.001$ & $+.004$ & $.837 (.010)$ & $+.005$ & $+.006$ \\
        CVPR & \hspace{1em}$+.011^{**}$ & $-.012$ & $.900 (.005)$ & $-.006$ & $-.006$ \\
        EMNLP & \hspace{1em}$+.022$ & $-.014$ & $.737 (.020)$ & $+.020$ & $+.018$ \\
        ICML & \hspace{1em}$+.006$ & $-.013$ & $.659 (.023)$ & $-.195^{*}$ & $-.233^{**}$ \\
        IJCAI & \hspace{1em}$-.013$ & $+.002$ & $.868 (.002)$ & $-.081^{**}$ & $-.067^{**}$ \\
        NAACL & \hspace{1em}$+.008$ & $-.001$ & $.757 (.019)$ & $+.017^{*}$ & $+.017$ \\
        NeurIPS & \hspace{1em}$-.035$ & $-.017$ & $.586 (.035)$ & $-.221^{**}$ & $-.178^{**}$ \\ \midrule \bottomrule 
    \end{tabular}}
    \caption{\texttt{C} vs \texttt{W} classification using other combinations of writing features. The ``74 features'' column shows the mean and stdev of F1 scores, and other columns show F1 scores \textit{relative to} the ``74 features'' column. $*$ and $**$ indicate $p<.005$ and $p<.001$ respectively (Bonferroni corrected), both on 2-tailed $t-$test with $dof=10$.\\ Interpretation 1: Dropping the article length features and/or the highly correlated MATTR features almost never make significant difference on the classification performance. \\ Interpretation 2: The part-of-speech and voice features, like the other feature groups in Table \ref{tab:top_or_not_venue_classify}, support similar or slightly worse classification performance than the collection of 74 features.}
    \label{tab:top_or_not_features_ablation}
\end{table*}

\subsection{\texttt{C} vs \texttt{W} classification details}
\label{subsec:appendix-classification-detail}
Except from RoBERTa, the classifier models are implemented by scikit-learn \citep{scikit-learn}. We run 6-fold cross validation: In each rotation, we use 4,1,1 folds as train, dev, and test data with stratified splitting. We sweep through a collection of models, select the best model on the dev set, and record the F1 score on the test set.

Following are the models used for predicting the \texttt{C} vs \texttt{W} appearance:
\begin{itemize}[nosep]
    \item SVM (\texttt{LinearSVC}) with \{L1, L2\} loss
    \item Logistic Regressions, with max iteration \{100,200\} and $C=\{0.5, 1.0, 2.0\}$ regularization.
    \item ExtraTreesClassifier, with random state 0 and \{16, 32, 64, 128\} estimators.
    \item RandomForestClassifier, with random state 0 and \{50, 100, 200\} estimators.
    \item Gradient boosting, with random state 0 and maximum depths \{2,3,4,5\}
    \item Multiple Layers Perceptrons of various hidden sizes: [10], [20], [40], [80], [10,10], [20,20], and [40,40]
\end{itemize}
Considering different venues and the combinations of features, there are $210\times 6$ ``classification folds''. In each fold, every model except for MLP produces a ``feature importance'' score for each feature. When the MLP classifiers have the best performance (this happens in 194 out of 1,260 folds, i.e., $15.4\%$), we skip the feature importance scores. For each of the 210 classification settings, we average the feature importance scores across all folds. This allows us to rank the most important features.

The run time for classifying \texttt{C} vs. \texttt{W} on all venues is around 8 minutes on a desktop with M1 chip, using all writing features. The time is around the same for tf-idf features, which have similar dimensions since we set $d=100$.

\subsection{Pairwise classification details}
In this section of experiments, we use 74 writing features. For the content-based features, we use tf-idf with 100 dimensions, containing both the abstract and bodytext. In classification of both writing and tf-idf features, we use the default MLPClassifier of scikit-learn \citep{scikit-learn}. All classifications here are 5-fold cross validations.

The run time of pairwise classification between the venues is 1.5 minutes for writing features, and 30 minutes for tf-idf. The run time of pairwise classification between the categories of venues is 4 minutes for writing features, and 15 minutes for tf-idf, on a desktop machine.  

\subsection{Most indicative features for venues}
Table \ref{tab:important_features_full} includes the most indicative features, their Spearman R and ATE values for all venues.

\begin{table*}
    \centering
    \resizebox{.9\linewidth}{!}{
    \begin{tabular}{c l l l l}
    \toprule 
        \textbf{Venue} & \textbf{Features} & \textbf{Spearman R} & \textbf{ATE} & \textbf{Interpretation} \\ \midrule 
        \multirow{5}{*}{AAAI} & \texttt{n\_author} & $+0.25^{**}$ & $+0.03$ & \texttt{C} papers are larger \\
        & \texttt{sent\_lens\_bodytext\_mean} & $-0.16^{**}$ & $+0.02$ & Ambiguous \\
        & \texttt{grammar\_errors\_bodytext} & $-0.01$ & $-0.02$ & \texttt{W} papers are larger\\
        & \texttt{flesch\_kincaid\_grade\_level\_bodytext} & $-0.08$ & $+0.01$ & Ambiguous \\
        & \texttt{outbound\_citations\_per\_word} & $+0.24^{**}$ & $+56.06$ & \texttt{C} papers are larger \\ 
        \midrule 
        \multirow{5}{*}{ACL} & \texttt{flesch\_kincaid\_grade\_level\_bodytext} & $-0.05$ & $+0.05$ & Ambiguous\\
        & \texttt{grammar\_errors\_abstract} & $-0.09^{**}$ & $-0.01$ & \texttt{W} papers are larger\\
        & \texttt{surprisal\_abstract\_std} & $-0.01$ & $+0.00$ & Ambiguous \\
        & \texttt{title\_word\_length} & $-0.09^{**}$ & $-0.01$ & \texttt{W} papers are larger\\
        & \texttt{voice\_bodytext\_active} & $+0.09^{**}$ & $+0.15$ & \texttt{C} papers are larger\\ \midrule
        \multirow{5}{*}{COLING} & \texttt{title\_word\_length} & $-0.06$ & $-0.01$ & \texttt{W} papers are larger\\
        & \texttt{n\_author} & $-0.03$ & $-0.01$ & \texttt{W} papers are larger\\
        & \texttt{surprisal\_abstract\_std} & $+0.05$ & $+0.02$ & \texttt{C} papers are larger\\
        & \texttt{sent\_lens\_bodytext\_mean} & $-0.02$ & $-0.01$ & \texttt{W} papers are larger\\
        & \texttt{surprisal\_abstrat\_mean} & $+0.03$ & $-0.01$ & Ambituous \\
        \midrule 
        \multirow{5}{*}{CVPR} & \texttt{grammar\_errors\_bodytext} & $+0.03$ & $+0.00$ & Ambiguous\\
        & \texttt{abstract\_word\_counts} & $-0.06$ & $+0.00$ & Ambiguous\\
        & \texttt{lex\_mattr\_10\_bodytext} & $+0.04$ & $+2.71$ & \texttt{C} papers are larger \\
        & \texttt{n\_outbound\_citations} & $+0.08^{**}$ & $+0.00$ & \texttt{C} papers are larger \\
        & \texttt{surprisal\_bodytext\_mean} & $+0.10^{**}$ & $+0.00$ & \texttt{C} papers are larger \\
        \midrule 
        \multirow{5}{*}{EMNLP} & \texttt{outbound\_citations\_per\_word} & $-0.17^{**}$ & $+67.6$ & Ambiguous \\
        & \texttt{n\_author} & $-0.17^{**}$ & $-0.05$ & \texttt{W} papers are larger \\
        & \texttt{grammar\_errors\_abstract} & $-0.18^{**}$ & $+0.01$ & \texttt{W} papers are larger \\
        & \texttt{n\_outbound\_citations} & $-0.09$ & $+0.09$ & Ambiguous \\
        & \texttt{abstract\_word\_counts} & $-0.16^{**}$ & $+0.00$ & \texttt{W} papers are larger \\ \midrule
        \multirow{5}{*}{ICML} & \texttt{n\_outbound\_citations} & $-0.33^{**}$ & $-0.01$ & \texttt{W} papers are larger\\
        & \texttt{abstract\_word\_counts} & $-0.23^{**}$ & $+0.00$ & Ambiguous \\
        & \texttt{outbound\_citations\_per\_word} & $-0.27^{**}$ & $-81.1$ & \texttt{W} papers are larger\\
        & \texttt{grammar\_errors\_bodytext} & $-0.21^{**}$ & $+0.00^{**}$ & \texttt{W} papers are larger\\
        & \texttt{lex\_mattr\_10\_abstract} & $+0.00$ & $-0.43$ & Ambiguous \\
        \midrule 
        \multirow{5}{*}{IJCAI} & \texttt{n\_outbound\_citaitons} & $-0.55^{**}$ & $-0.06$ & \texttt{W} papers are larger\\
        & \texttt{outbound\_citations\_per\_word} & $-0.50^{**}$ & $-265$ & \texttt{W} papers are larger\\
        & \texttt{n\_author} & $+0.26^{**}$ & $+0.06$ & \texttt{C} papers are larger\\
        & \texttt{grammar\_errors\_bodytext} & $-0.26^{**}$ & $+0.00$ & Ambiguous \\
        & \texttt{lex\_mattr\_10\_bodytext} & $+0.14^{**}$ & $+0.86$ & \texttt{C} papers are larger \\
        \midrule 
        \multirow{5}{*}{NAACL} & \texttt{abstract\_word\_counts} & $-0.20^{**}$ & $+0.00$ & \texttt{C} papers are larger\\
        & \texttt{grammar\_errors\_bodytext} & $-0.15^{**}$ & $+0.00$ & \texttt{W} papers are larger\\
        & \texttt{n\_author} & $+0.01$ & $+0.01$ & \texttt{C} papers are larger\\
        & \texttt{surprisal\_bodytext\_mean} & $+0.00$ & $+0.01$ & Ambiguous \\
        & \texttt{surprisal\_bodytext\_std} & $-0.07^{**}$ & $+0.11$ & Ambiguous \\
        \midrule 
        \multirow{5}{*}{NeurIPS} & \texttt{flesch\_kincaid\_grade\_level\_bodytext} & $+0.05$ & $-0.01$ & Ambiguous \\
        & \texttt{n\_author} & $-0.10^{**}$ & $-0.01$ & \texttt{W} papers are larger\\
        & \texttt{surprisal\_bodytext\_std} & $+0.06$ & $+0.00$ & \texttt{C} papers are larger \\
        & \texttt{abstract\_sent\_counts} & $-0.19^{**}$ & $-0.09$ & \texttt{W} papers are larger\\
        & \texttt{rst\_Elaboration} & $-0.01$ & $-0.64$ & \texttt{W} papers are larger\\ \midrule
        \bottomrule
    \end{tabular}}
    \caption{The most important 5 writing features for classifying \texttt{C} vs. \texttt{W} appearance, their Spearman R, and their estimated average treatment effects (ATE). $*$ and $**$ indicate $p<.005$ and $p<.001$ respectively (Bonferroni corrected). For Spearman R, the $p$ value indicates the likelihood the feature and target come from identical distributions. For ATE, the $p$ value is computed by \textit{doWhy}'s default bootstrapping test.}
    \label{tab:important_features_full}
\end{table*}

\end{document}